\documentclass[journal,twoside,web]{ieeecolor}
\usepackage{tmi}
\usepackage{cite}
\usepackage{amsmath,amssymb,amsfonts}
\usepackage{algorithmic}
\usepackage{graphicx}
\usepackage{textcomp}

\usepackage{graphicx}
\usepackage{array}
\usepackage{subfigure}
\usepackage{setspace}
\usepackage{amssymb}
\usepackage{booktabs}
\usepackage{stfloats}
\usepackage[justification=centering]{caption}
\usepackage{amsmath}
\usepackage{autobreak}
\usepackage[colorlinks,linkcolor=blue,anchorcolor=blue,citecolor=blue]{hyperref}
\usepackage{bbm}
\usepackage{threeparttable}

\usepackage{booktabs,multirow}
\usepackage{multicol} 
\usepackage{mwe}

\usepackage[linesnumbered,boxed,ruled,commentsnumbered]{algorithm2e}

\def\BibTeX{{\rm B\kern-.05em{\sc i\kern-.025em b}\kern-.08em
    T\kern-.1667em\lower.7ex\hbox{E}\kern-.125emX}}
\begin{document}

\title{Domain Generalization on Medical Imaging Classification using Episodic Training with Task Augmentation} 
\author{Chenxin Li, Qi Qi, Xinghao Ding, Yue Huang, Dong Liang, 
and Yizhou Yu, \IEEEmembership{Fellow, IEEE}  
\thanks{This work was partially supported by National Key Research and Development Program of China (No.2020YFC2003900) and the National Natural Science Foundation of China under Grants U19B2031, 61971369, in part by Fundamental Research Funds for the Central Universities 20720200003.}
\thanks{Chenxin Li, Qi Qi, Xinghao Ding, Yue Huang are with the School of Informatics, Xiamen University, Xiamen 361005, China. (e-mail:huangyue05@gmail)}    
\thanks{Dong Liang is with the Paul C. Lauterbur Research Center for Biomedical Imaging, Shenzhen Institutes of Advanced Technology, Chinese Academy of Sciences, Shenzhen 518055, China. (dong.liang@siat.ac.cn)}
\thanks{Yizhou Yu are with the Deepwise AI Laboratory, Beijing 100125, China.
(e-mail: yizhouy@acm.org)}
}

\maketitle

\begin{abstract}
Medical imaging datasets usually exhibit domain shift due to the variations of scanner vendors, imaging protocols, etc.
This raises the concern about the generalization capacity of machine learning models.
Domain generalization (DG), which aims to learn a model from multiple source domains such that it can be directly generalized to unseen test domains, seems particularly promising to medical imaging community. 
To address DG, recent model-agnostic meta-learning (MAML) has been introduced, which transfers the knowledge from previous training tasks to facilitate the learning of novel testing tasks.
However, in clinical practice, there are usually only a few annotated source domains available,
which decreases the capacity of training task generation and thus increases the risk of overfitting to training tasks in the paradigm.
In this paper, we propose a novel DG scheme of episodic training with task augmentation on medical imaging classification.
Based on meta-learning, we develop the paradigm of episodic training to construct the knowledge transfer from episodic training-task simulation to the real testing task of DG.
Motivated by the limited number of source domains in real-world medical deployment, we consider the unique task-level overfitting
and we propose task augmentation to enhance the variety during training task generation to alleviate it.
With the established learning framework, we further exploit a novel meta-objective to regularize the deep embedding of training domains.
To validate the effectiveness of the proposed method, we perform experiments on histopathological images and abdominal CT images.

\end{abstract}

\begin{IEEEkeywords}
Medical imaging classification, domain generalization, model-agnostic meta-learning, episodic training, task-level overfitting
\end{IEEEkeywords}

\IEEEpeerreviewmaketitle

\section{Introduction}
\IEEEPARstart{M}{edical} imaging analysis usually provides crucial clinical information for diagnosis.
For instance, in oncology studies, the classification of epithelium and stroma for histopathological images is a prior step to calculate epithelium-stroma (ES) ratio. Then this ratio can be used to perform patient stratification in many different tumor types, especially breast cancer \cite{downey2014prognostic}.
Another example is abdominal computed tomography (CT) images. The accurate segmentation of objective organ from the surrounding regions is crucial for the radiation therapy planning and follow-up evaluation \cite{heimann2009comparison}.

In clinical practice, manual annotations are tedious, error-prone and time-consuming.
They are even more difficult to acquire in some conditions where specialized biomedical expert knowledge is required. 
The computer-assisted automatic algorithms are thus developed, including the classical machine learning methods \cite{erickson2017machine} and recent deep convolutional neural networks (CNN) \cite{shen2017deep}.
However, these methods are based on IID assumption, i.e., {\itshape training and testing set are independent and identically distributed} \cite{IID}, and they would suffer performance degradation when IID is not satisfied \cite{non-IID1}.
Moreover, this non-IID scenario (also called as domain shift problem) is common in clinical practice, where the medical images are usually acquired by different institutions with various types of scanner vendors, patient populations, disease severity, etc. 
To cope with it, a straightforward way is to fine-tune \cite{finetune1}, but it requires extra annotated data from the test domain.  
Without requiring the annotation on test domain, unsupervised domain adaptation (UDA) \cite{uda1,uda2,uda3,sifa} seems more appealing.
However, it still needs to access the test-domain data during the training procedure.
Then, for every new test domain, we have to repeatedly collect the training images from the test domain, which limits real-world medical deployment.
\begin{figure}[!t]   
	\centering	   
	\includegraphics[width=\linewidth,height=5.3cm]{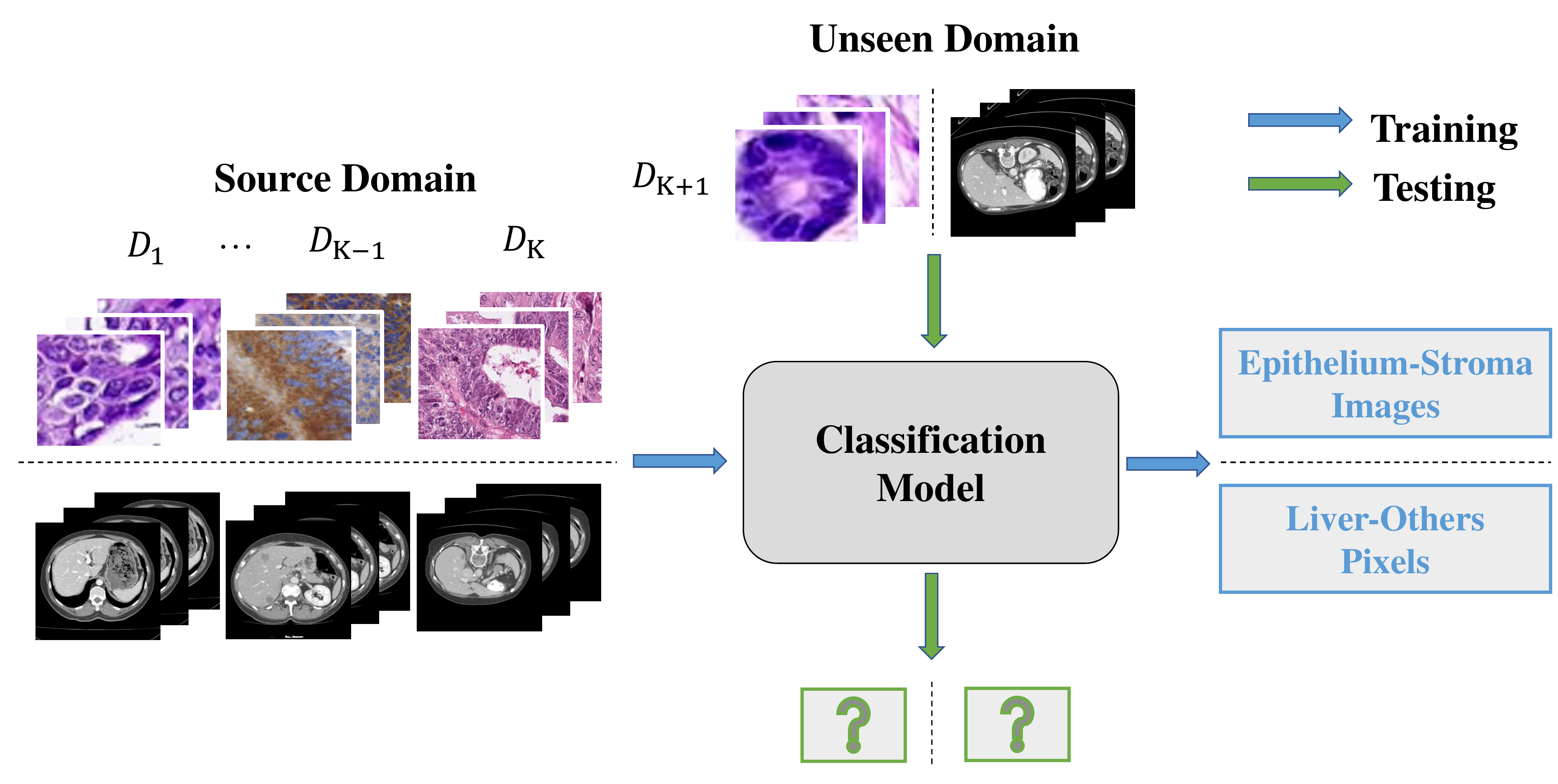}         
	\caption{Pipeline of domain generalization (DG) on medical images. Trained on multiple annotated source domains $\{\mathcal{D}_1, \mathcal{D}_2, . . . , \mathcal{D}_K\}$, the model can directly generalize to an unseen test domain $\mathcal{D}_{K+1}$, i.e., without accessing any information of $\mathcal{D}_{K+1}$ during the training phase.
	 } \label{fig:DG}      
\end{figure}

In contrast, it has a broad prospect for clinical practice to train a generalizable model that can be directly applied to a new domain (cf. Figure \ref{fig:DG}), i.e., domain generalization (DG) \cite{li2018domain,jigsaw,bigaug,yoon2019generalizable,masf,liu2021feddg}. 
Without accessing the information of test domain, the principle of DG is to capture the representation that is insensitive to the changes of domain-specific statistics.
To achieve this goal, some works explore data augmentation or self-supervised learning signals \cite{jigsaw,bigaug} to promote the model to learn a domain-invariant embedding.
However, these designs are somehow heuristic and highly rely on the characteristics of tasks.
The others aim to align the feature embedding of multiple source domains \cite{li2018domain,li2018deep}, which has shown effectiveness on domain adaptation (DA). 
However, embedding alignment between source domains does not necessarily lead to the alignment between source and unseen test domains.  

Recently, model-agnostic meta-learning (MAML) \cite{maml}, which aims to train the model that can be fast adapted to new tasks, has been promising in many applications like few-shot learning.
Then this spirit is introduced to address DG with a paradigm of episodic training \cite{yoon2019generalizable,masf,liu2021feddg}.
A key difference between episodic training and conventional supervised training is that the data point in episodic training is a task.
Specifically, a meta-task (virtual DG task) is generated episodically, with the domain shift simulation from splitting the available source domains into meta-train (virtual training) and meta-test (virtual testing) domains.
This scheme implies the knowledge transfer of coping with domain shift from virtual training tasks to real testing tasks, based on the task-level similarity between them. 


However, since the test domain is unseen during training, the real domain shift is also unobserved. Thus it inevitably exists a shift between the training task with simulated domain shift and the testing task with real domain shift.
Furthermore, 
in clinical practice, it is common that there are only several labeled source domains available. 
This makes the meta-task generation as well as domain shift simulation restricted into just a few patterns, which further limits the capacity of training task generation. 
Thus, the risk of overfitting to training tasks in the episodic paradigm is increased,
under which the model would suffer the performance drop of DG from simulated training task to real testing task.

In this work, we aim to alleviate this special risk of task-level overfitting.
We propose task augmentation to enhance the variety for the meta-task generation, which promotes the model to extract the task-invariant knowledge against various domain shift simulations.
We further incorporate a novel meta-update objective in the proposed episodic learning framework to perform the regularization for deep semantic embedding.

Our main contributions are summarized as follows.
\begin{itemize}
\item We propose a novel DG method for generalizable medical image classification on unseen test domains, through the paradigm of episodic training.
\item We consider the risk of overfitting to the simulated training tasks in the existing scheme of episodic training, which is motivated by the limited number of available annotated source domains in clinical practice.
\item We develop the task augmentation to alleviate this unique task-level overfitting, using our proposed mixed task sampling (MTS) strategy to enhance the variety for the training task simulation.
\item We exploit the proposed learning framework in the deep semantic embedding with a novel meta-objective, which regularizes the cross-domain alignment from both sample-wise and prototype-wise.
\item We conduct extensive experiments on two typical tasks, including
epithelium-stroma classification on histopathological images, and liver segmentation on abdominal CT images.
The results demonstrate the superiority of our method over the state-of-the-art DG methods.
\end{itemize} 

\section{Related Work}
In this section, we review medical imaging classification, mainly on the practice of two typical tasks. Then we survey some methods of domain adaptation (DA). Finally, we review the recent domain generalization (DG) methods.

\subsection{Medical Imaging Classification}
Medical imaging classification is critical in many clinical applications \cite{shen2017deep}. In this paper, we investigate two common tasks, i.e., epithelium-stroma classification and liver segmentation (which can be treated as pixel-wise classification).
In oncology studies, epithelium and stroma are basic animal tissue types in histopathological images.
The ratio of epithelium-stroma (ES) can be used to perform patient stratification and follow-up \cite{downey2014prognostic}, and the classification of epithelium-stroma is a prior step for calculating it.
In the early machine learning methods, hand-crafted features are extracted according to the characteristics of histopathological images \cite{bianconi2015discrimination, nava2015classification}. 
Recently, deep learning approaches achieve remarkable performance \cite{huang2017epithelium1,qi2018label}, and some methods further investigate the generalization ability of deep models on epithelium-stroma classification \cite{huang2017epithelium2,qi2020curriculum}.

Moreover, in computer-aided hepatic disease diagnosis and treatment planning, the accurate segmentation of liver organ is a crucial prerequisite. 
Statistical deformable models utilize shape priors, intensity distributions as well as boundary and region information \cite{kainmuller2007shape,wimmer2009generic}. 
Learning-based methods have also been explored to seek powerful features \cite{al2015automatic}.
Recently, convolutional neural networks (CNNs) have been widely utilized in the liver segmentation \cite{dou20163d,yang2017automatic}. But the variety of imaging protocols, scanner etc. further raises the concern on generalization ability of these models. 

\subsection{Domain Adaptation} 
Domain adaptation (DA) plays a crucial role in cross-dataset evaluation especially when the training and testing set have domain shift. 
Generally, it is performed by mitigating the distribution mismatch between the source and target domains\cite{long2016unsupervised}.
In medical image analysis, many works focus on feature distribution alignment by minimizing the distance between the feature distributions of source and target domain \cite{uda1,uda2,uda3,sifa}, using maximum mean discrepancy, adversarial learning, or regularization term. 
Furthermore, some methods consider to conduct the domain alignment and refine the task-specific decision boundary at the same time \cite{saito2018maximum,zhu2019aligning}. This motivates us to incorporate the paradigm of episodic learning in the deep embedding alignment from both  sample-wise and prototype-wise perspectives.

\subsection{Domain Generalization} 
As a recently emerged concept, domain generalization (DG) aims to learn a model from multiple distributed source domains that can generalize to unseen test domains. 
To achieve this, some methods try to learn domain-invariant representation by minimizing the domain discrepancy across source domains, using maximum mean discrepancy (MMD) \cite{muandet2013domain}, contrastive loss \cite{8237871}, adversarial learning \cite{li2018deep, li2018domain}, autoencoders \cite{ilse2019diva}.    
The others design heuristic schemes, such as  
manipulating deep neural network architectures \cite{li2017deeper}, leveraging self-supervision signals \cite{jigsaw}, designing novel training paradigm \cite{li2019episodic,wang2020dofe}, or conducting data augmentation \cite{volpi2018generalizing,bigaug}.
Very recently, with the rise of model-agnostic meta-learning (MAML) \cite{maml}, an paradigm of episodic learning is introduced to address DG \cite{li2018learning, balaji2018metareg, li2019feature,masf,liu2021feddg}. 
The principle behind this is to simulate virtual meta-task episodically so that the knowledge of maintaining generalization under domain shift can be transferred to the real task of DG.
However, there is still a shift between virtual training task and real testing task, which raises the risk of overfitting to training tasks in the paradigm. 



\begin{figure*}[!t]  
	\centering	   
	\includegraphics[width=\textwidth,keepaspectratio]{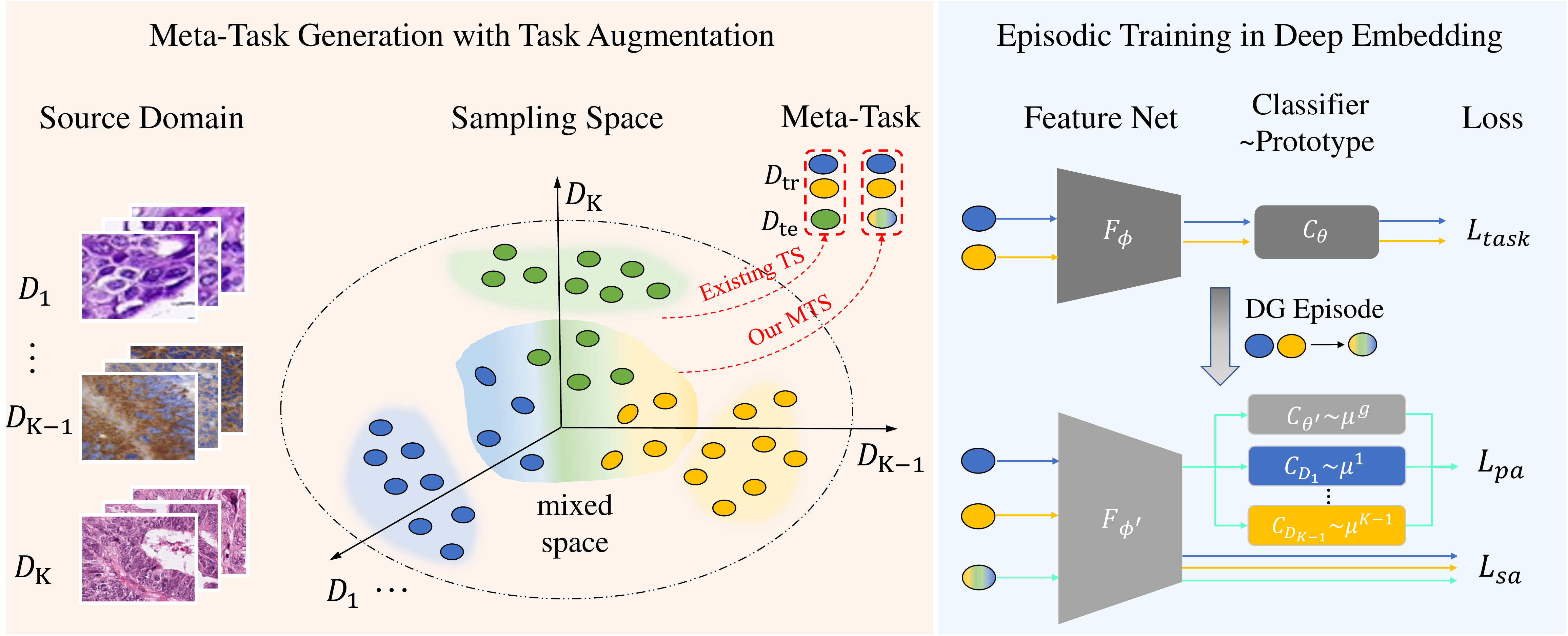}   
	\caption{Overview of the proposed domain generalization (DG) method using episodic training with task augmentation.
	The meta-task is simulated from training domains episodically as a virtual training task.
	The task augmentation is performed to enhance the variety of meta-task using mixed task sampling (MTS).
	The overall proposed training framework is incorporated in the cross-domain deep embedded alignment from sample-wise $\mathcal{L}_{sa}$ and prototype\protect\footnotemark-wise $\mathcal{L}_{pa}$.} \label{fig:pipeline}       
\end{figure*} 
\footnotetext{Prototype \cite{snell2017prototypical,chen2019closer} means the centroid of the embedding for same-class samples. It is usually used in conjunction with cosine distance to perform a decision boundary or classifier.}

\section{Method}
In this paper, following the common setting of domain generalization (DG) \cite{gidaris2018dynamic}, we denote input and label space by $X$ and $Y$, and the $K$ source domains available during training as $\{\mathcal{D}_1, \mathcal{D}_2, . . . , \mathcal{D}_K\}$ with different distributions on the joint space $X \times Y$.
$Y$ is the discrete set $\{1, 2, . . . C\}$ shared by all domains and $C$ denotes the number of classes. Samples are drawn from the domain dataset $\mathcal{D}_k = \{(x_{n},y_n) \}_{n=1}^{N_k}$, where $N_k$ is the number of samples in $k$ domain.
The goal of DG is to use multiple source domains $\{\mathcal{D}_1, \mathcal{D}_2, . . . , \mathcal{D}_K\}$ to learn a mapping $X \rightarrow Y$ that can generalize well on an unseen test domain $\mathcal{D}_{K+1}$,
i.e., without accessing any information of $\mathcal{D}_{K+1}$ during the training phase.

The overview of the proposed method is presented in Fig. \ref{fig:pipeline}. The remainders of this section are organized as follows. 
First, we describe the model-agnostic episodic training backbone.
Then, we introduce task augmentation, which is proposed to enhance task-level variety via a mixed task sampling (MTS) strategy. 
Finally, we present the meta-objective to regularize the proposed learning procedure in the deep semantic embedding.

\subsection{Episodic Training Backbone for DG} \label{para1}
Generally, we consider a basic model in medical image classification, consisting of a feature extractor $\mathcal{F}_\phi:X \rightarrow Z$ and a classifier network $\mathcal{C}_\theta:Z\rightarrow Y$, where $Z$ is the lower-dimensional feature space.\footnote{For segmentation task, Z is the high-dimensional space for feature maps.} 
In standard supervised learning, the parameters $(\phi, \theta)$ are usually optimized on a task-specific loss $\mathcal{L}_{task}$, e.g. cross-entropy loss. 
Although the minimization of $\mathcal{L}_{task}$ may produce highly discriminative features $z = \mathcal{F}_{\phi}(x)$ as well as a well-behaved classifier on training domains, nothing in this process emphasizes the generalization performance of the deep model to unseen test domains.
In contrast, based on model-agnostic meta-learning (MAML) \cite{maml}, the paradigm of episodic training optimizes the model according to the performance on the virtual test domain, which reflects the generalization ability.


In particular, the virtual training task of DG, i.e., meta-task $\mathcal{T}_i$  
is episodically sampled from the set ${\mathcal{T}=\{\mathcal{T}_1,\mathcal{T}_2,...,\mathcal{T}_I}\}$, and the available source domains are split into meta-train $\mathcal{D}^{\mathcal{T}_i}_{tr}$ and meta-test $\mathcal{D}^{\mathcal{T}_i}_{te}$ domains that depends on $\mathcal{T}_i$.   
Then two loops of optimization are applied on the model training procedure.
The first one is {\itshape inner optimization}, where the model of feature net $\mathcal{F}_\phi$ and classifier $\mathcal{C}_\theta$ is updated with one or more steps of gradient descent on $\mathcal{D}^{\mathcal{T}_i}_{tr}$ with respect to a task loss $\mathcal{L}_{task}$:
\begin{equation} 
\label{eq_inner}
(\phi'_i,\theta'_i) \leftarrow (\phi,\theta) - \alpha \nabla_{( \phi, \theta)} \mathcal{L}_{task}(D^{\mathcal{T}_i}_{tr}; \phi, \theta)
\end{equation} 
where $\alpha$ denotes the learning rate w.r.t. the inner optimization. 
This leads to a temporary predictive model $\mathcal{C}_{\theta'_i}\circ \mathcal{F}_{\phi'_i}$ with the high task accuracy on the meta-train domains $\mathcal{D}^{\mathcal{T}_i}_{tr}$.
To further evaluate the generalization performance,
the second {\itshape outer optimization} is conducted on the generated temporary model with weights of $(\phi'_i,\theta'_i)$. Specifically, besides the task-specific loss, a meta-learning objective is utilized to emphasize some certain properties that we desire the temporary predictive model to exhibit. Specially, the meta-objective is calculated on the temporary model with weights of $(\phi'_i,\theta'_i)$ while the gradients are computed towards the origin model with weights of$(\phi,\theta)$:
\begin{equation}
\label{eq_outer}
\begin{aligned}
( \phi, \theta) \leftarrow ( \phi, \theta) - \beta \nabla_{( \phi, \theta)}\{&\mathcal{L}_{task}(D^{\mathcal{T}_i}_{tr}; \phi, \theta)\\
&+\mathcal{L}_{meta}(D^{\mathcal{T}_i}_{tr}, D^{\mathcal{T}_i}_{te}; \phi'_i, \theta'_i)\},
\end{aligned}
\end{equation}
where $\beta$ is the learning rate w.r.t. outer optimization. 

\subsection{Task Augmentation via Mixed Task Sampling}

The paradigm of episodic learning optimizes the model to behave well on the virtual training task of DG such that it can also behave well on the real testing task, based on the similarity between training and testing tasks. 
However, the risk of overfitting to training tasks emerges due to the inevitable shift between training and testing tasks.
This risk could be increased in clinical practice where it is often seen that only several labeled source domains are available,
which further limits the capacity of training task generation.
To alleviate this special overfitting, inspired by the art of data augmentation, we propose the novel task augmentation to enhance the variety during training meta-task simulation.

Specifically, in the existing paradigm of episodic training, the meta-task is generated by the task sampling (TS) strategy that $K$ available source domains are split into $K-1$ meta-train and the held-out meta-test to simulate domain shift. 
Intuitively, the number of meta-task patterns is $K$ (i.e., $C_K^1$).
This is limited when $K$ is small (e.g., $K=3$), which is a common phenomenon in real-world medical deployment.
In comparison, we perform the task augmentation to enhance the capacity of meta-task simualtion. We present the mixed task sampling (MTS) strategy,
where the meta-train domains are generated in the same way while the meta-test domain is acquired by the interpolation among all the source domains:
\begin{equation}
\label{eq_Dte}
\mathcal{D}_{te} =\bigcup_{k=1}^{K}  Sample(\mathcal{D}_k, \; r_k\cdot N_{te}), \quad  s.t. \sum^K_{k=1} r_k = 1 
\end{equation}
where $Sample(\mathcal{D},N)$ is the sampling operator that we define to randomly select $N$ samples from domain $\mathcal{D}$.
$N_{te}$ denotes the number of samples for $D_{te}$ and $r_k$ represents the interpolation ratio for $k$-$th$ source domain.
In particular, the existing dominant TS strategy is the special case of our MTS approach when $r_K=1$.
Furthermore, compared to the previous TS approach, the proposed MTS increases the capacity of meta-task patterns largely from original $K$,
which highlights the variety enhancement during training task generation.

\subsection{Meta-Objective to Regularize Deep Semantic Embedding}
Here we further discuss the design of meta-learning objective $\mathcal{L}_{meta}$ in Eq. \ref{eq_outer}. 
A straightforward approach is to encourage the high task accuracy on meta-test domain, i.e., $\mathcal{L}_{meta}(D^{\mathcal{T}_i}_{tr}, D^{\mathcal{T}_i}_{te}; \phi'_i, \theta'_i)=\mathcal{L}_{task}(D^{\mathcal{T}_i}_{te}; \phi'_i, \theta'_i)$, which exhibits the generalization performance of the model.
Recently, the alignment of semantic embedding across training domains has been developed in many tasks like DA and DG \cite{uda1,uda2,li2018domain,li2018deep,masf,liu2021feddg}, which promotes the model to learn class-discriminative whilst domain-invariant representation.
In light of this, we incorporate the meta-objective of episodic learning in the learned deep embedding, with the regularization of cross-domain alignment from sample-wise and prototype-wise, respectively.

Specially, we use sample-wise alignment to encourage clustering the features of the same class while keeping the features of different class far apart.
Equipped with contrastive learning \cite{hadsell2006dimensionality} and cosine distance based decision boundary \cite{chen2019closer,snell2017prototypical}, the loss of sample-wise alignment is:

\begin{small}
\begin{equation}
\label{eq_fa}  
\begin{aligned}
&\mathcal{L}_{sa}( D^{\mathcal{T}_i}_{tr}, D^{\mathcal{T}_i}_{te}; \phi'_i, \theta'_i)=\\
&\sum_{x_j^c\in(D^{\mathcal{T}_i}_{tr},D^{\mathcal{T}_i}_{te})} \{ (1-\cos(\mathcal{F}(x_j^c), \mu^g_c)) + \sum_{d=1}^C \mathbbm{1}(d \neq c) \cos(\mathcal{F}(x_j^c), \mu^g_d) \}
\end{aligned}
\end{equation}
\end{small}
where $\mathbbm{1}(\cdot)$ is an indicator function. $\mu^g = [\mu^g_1,\mu^g_2,...,\mu^g_C]$ denotes the set of domain-general prototypes. They are actually the weight vectors of classifier $\mathcal{C}$, as $\cos(\mathcal{F}(x_j^c),\mu^g_c)=\frac{\mathcal{F}(x_j^c)}{||\mathcal{F}(x_j^c)||} \cdot \frac{\mu^g_c}{||\mu^g_c||} $.
Moreover, we introduce the domain-specific prototypes, which are dependent on the respective statistics of training domains.
Specifically, for domain $k$, we calculate the centroid of class $c$ as $\mu^k_c=\frac{1}{n_c^k}\sum_{( x_j^k ,y_j^k = c )\in D_k} \mathcal{F}(x_j^k)$, and the prototypes of all classes are 
$\mu^k=[\mu^k_1,\mu^k_2,...,\mu^k_C]$.
With these prototypes, the prediction scores of an input sample can be obtained by calculating the cosine distance between feature embedding and prototypes.
In particular, in the meta-task $\mathcal{T}_i$, assume that the meta-train is composed of the $K-1$ domains $\{\mathcal{D}_1,\mathcal{D}_2,...,\mathcal{D}_{K-1}\}$ in all $K$ source domains. 
Given a query sample $x_j$ from meta-test, we calculate K prediction scores in total based on the $K-1$ domain-specific prototypes $\{\mu^{1},\mu^{2},...,\mu^{{K-1}}\}$ and the domain-general prototypes $\mu^g$. We denote the prediction scores for sample $x_j$ as $\{P^{1}_j,P^{2}_j,...,P^{{K-1}}_j,P^g_j\}$.
Then the prototype-wise alignment is performed to encourage the prediction scores across different prototypes to be consistent with each other, i.e., the loss of prototype-wise alignment is:

\begin{small}
\begin{equation}
\label{eq_ca}
\mathcal{L}_{pa}( D^{\mathcal{T}_i}_{tr}, D^{\mathcal{T}_i}_{te}; \phi'_i, \theta'_i) =\sum_{a \neq b}\sum_{x_j \in D_{te}} \frac{1}{2}(KL(P_j^a||P_j^b)+KL(P_j^b||P_j^a))
\end{equation}
\end{small}
where 
\begin{footnotesize}
$(a,b) \in \{1,2,...,K-1, g\}$, $KL(P_j^a||P_j^b) = \sum_{c=1}^C P^a_{jc}*\log{\frac{P^a_{jc}}{P^b_{kc}}}$.
\end{footnotesize}
Finally, the overall meta-objective is used to regularize the proposed episodic training framework in the deep embedding by integrating the cross-domain alignment from both sample-wise and prototype-wise: 
\begin{equation}
\label{eq_meta}
\begin{aligned}
\mathcal{L}_{meta}( D^{\mathcal{T}_i}_{tr}, D^{\mathcal{T}_i}_{tr}; \phi'_i, \theta'_i) = \gamma_1 \mathcal{L}_{fa} + \gamma_2\mathcal{L}_{ca}
\end{aligned}
\end{equation}
where $\gamma_1$ and $\gamma_2$ are the trade-off parameters to balance the two alignment losses. The algorithm is shown in Algorithm \ref{al}.

\IncMargin{1em} 
\begin{algorithm}

    \SetAlgoNoLine
    \SetKwInOut{Input}{\textbf{Input}}\SetKwInOut{Output}{\textbf{Output}}
    \Input{
		Source training domains $\{D_k\}^K_{k=1}$; \\
        Hyperparameters $\alpha, \beta, \gamma_1, \gamma_2 > 0$;\\
        Maximum number of iterations $I$;\\
        }
    \Output{
        Feature extractor $\mathcal{F}_{\phi}$, classifier network $\mathcal{C}_{\theta}$,
        }
    \BlankLine
    \For{$ i = 1 : I$}{
    Generate a meta-task $\mathcal{T}_i$, with meta-train $\mathcal{D}_{tr}$ and meta-test $\mathcal{D}_{te}$ sampled from source domains $\mathcal{D}$ via MTS \tcp*{Eq.\ref{eq_Dte}}
    $(\phi'_i,\theta'_i) \leftarrow  (\phi,\theta) - \alpha \nabla_{( \phi, \theta)} \mathcal{L}_{task}(D^{\mathcal{T}_i}_{tr}; \phi, \theta)$ \tcp*{{\itshape Inner optim}, Eq.\ref{eq_inner}}
    Compute sample-wise alignment loss: $\mathcal{L}_{sa}( D^{\mathcal{T}_i}_{tr}, D^{\mathcal{T}_i}_{te}; \phi'_i, \theta'_i)$ \tcp*{Eq.\ref{eq_fa}}
    Compute prototype-wise alignment loss: 
	$\mathcal{L}_{pa}(D^{\mathcal{T}_i}_{tr}, D^{\mathcal{T}_i}_{te}; \phi'_i, \theta'_i)$ \tcp*{Eq.\ref{eq_ca}}
    $\mathcal{L}_{meta} = \gamma_1 * \mathcal{L}_{sa} + \gamma_2 * \mathcal{L}_{pa}$ \tcp*{Eq.\ref{eq_meta}}
    $( \phi, \theta) \leftarrow ( \phi, \theta) - \beta \nabla_{( \phi, \theta)}(\mathcal{L}_{task}+\mathcal{L}_{meta})$ \tcp*{{\itshape Outer optim}, Eq.\ref{eq_outer}}
	}
    \caption{The proposed paradigm of episodic training with task augmentation for DG.\label{al}}
\end{algorithm}
\DecMargin{1em}

\begin{table*}[!t]
	\centering
	\caption{Statistics of the datasets used in our experiments.}
	\resizebox{0.92\linewidth}{!}{
	  \begin{tabular}{cccccc|ccc}
	  \toprule
	  \multicolumn{6}{c|}{Histopathological Images} & \multicolumn{3}{c}{Abdominal CT Scans} \\
	  \hline
	  Dataset & \multicolumn{1}{c}{Staining method} & \multicolumn{1}{c}{Organ} & \multicolumn{1}{c}{Scanner} & \multicolumn{1}{c}{Location} & \multicolumn{1}{c|}{\# Image} & Dataset & \multicolumn{1}{c}{\# Subject} & \multicolumn{1}{c}{\# Image} \\
	  \hline
	  VGH   &  H\&E &  Breast        &  Aperio     &   Netherlands    & 5920  & BTCV  &  30    & 5488   \\
	  \hline
	  NKI   &  H\&E &  Breast        &  Aperio     &   Vancouver    & 8337  & CHAOS &    20  &   2341\\
	  \hline
	  IHC   &  IHC  &  Coloretal     &  Mirax Scan  &   Finland     & 1376 & IRCAD &     20  &    2074\\
	  \hline
	  NCH   &  H\&E &  Coloretal     &  Unknown     &   Germany   & 26417 & LITS  &       40  &    5785\\
	  \bottomrule
	  \end{tabular}%
	}
	\label{tab:dataset}%
  \end{table*}%

\section{Experimental Results}

\subsection{Dataset and Evaluation Metrics}

\begin{figure}[!t]
	\centering	 
	\includegraphics[width=\linewidth,keepaspectratio]{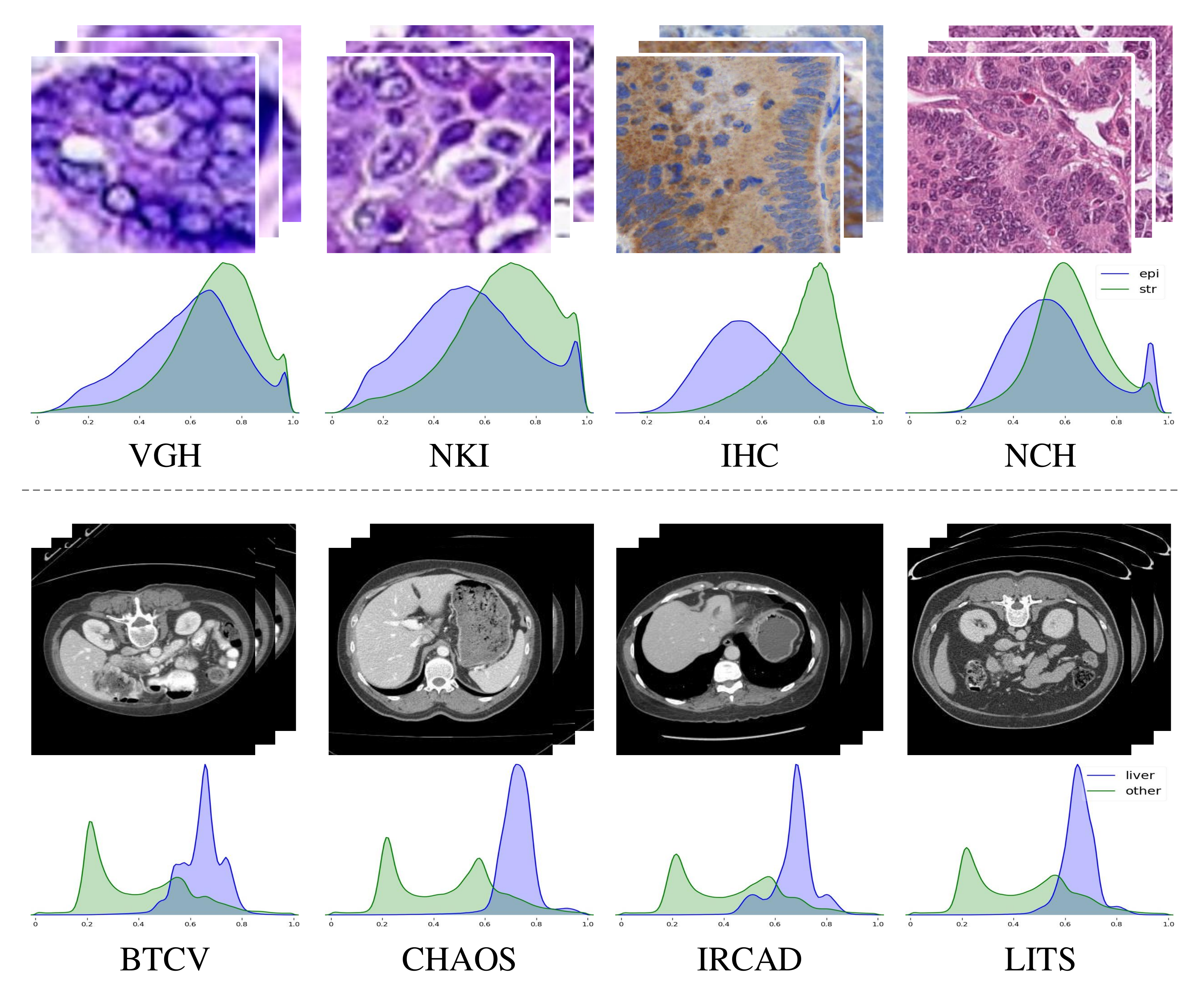}
	\caption{Histopathological images and abdominal CT scans from diverse datasets. The example images and the intensity histograms for every dataset are present.} \label{fig:dataset}
\end{figure}

We evaluate the proposed method on epithelium-stroma classification, using the histopathological images from four common publicly available datasets, including VGH \cite{VGH_NKI}, NKI \cite{VGH_NKI}, IHC \cite{IHC}, and NCH \cite{NCH}. 
Among them, each image is labeled as the class of epithelium or stroma. 
They are collected from different staining methods, human organs, scanners as well as institutions, which causes the domain shift among them. 
The detailed information is shown in Figure \ref{fig:dataset} and Table \ref{tab:dataset}.
The input images are resize to 224$\times$224, 
following previous practice \cite{huang2017epithelium1,huang2017epithelium2,qi2018label,qi2020curriculum}.

We also validate our method on the abdominal CT images from four public data sources, including BTCV \cite{BTCV}, CHAOS \cite{CHAOS}, IRCAD \cite{IRCAD}, LiTS \cite{LiTS}. 
Although some of these datasets are labeled in a multi-organ manner, the organ that is commonly labeled in the four datasets is only liver. So we conduct DG of liver segmentation task on these datasets.
They exhibit heterogeneous distributions due to the varying imaging conditions such as different scanners, clinical centers and subjects. 
The detailed information of each dataset is shown in Figure \ref{fig:dataset} and Table \ref{tab:dataset}.
Please note the details of data acquisition for the abdominal CT datasets are not present because they are not provided in their official dataset descriptions. 
Meanwhile, we use only the first forty subjects in LiTS such that the number of used subjects is in a similar scale compared to other datasets.
For pre-processing, the HU-value of CT scans are truncated to the range $[-200,250]$ to remove the irrelevant information, and then they are normalized to $[0,1]$ in intensity value. 
All the images are resized to 128$\times$128 in the axial plane. 
Simple data augmentation as random flipping is applied. 

Following the common practice in DG \cite{jigsaw,masf,liu2021feddg}, we use leave-one-domain-out cross-validation, i.e., training on multiple source domains and testing on the remaining unseen one, with randomly dividing the samples of each domain into 70\% training data and 30\% testing data.
To evaluate the performance on epithelium-stroma classification, we adopt the accuracy metric as $ACC = \frac{N_{c}}{N_{t}}$, where $N_c$ and $N_t$ represents the number of correctly classified images and total number of images, respectively. To evaluate the performance on liver segmentation, we adopt the most commonly-used metric, i.e., Dice coefficient (DC). It is calculated as $DC=\frac{{2|P\cap G|}}{|P|+|G|}$, where P and G denote the set of pixel-wise binary prediction and ground-truth, respectively.

\subsection{Implementation Details}
For image classification of epithelium-stroma, we adopt the AlexNet \cite{krizhevsky2012imagenet} pre-trained on ImageNet as the model backbone. The feature extractor $\mathcal{F}$ comprises the top layer of the AlexNet up to the pool7 layer and an additional bottleneck layer fc8 with 512 units. Accordingly, the classifier $\mathcal{C}$ is composed of the fully connected layer with weight size as $512\times2$. 
For the trade-off parameters in $\mathcal{L}_{fa}$ and $\mathcal{L}_{ca}$, we set $\gamma_1 = 1.0$ and $\gamma_2 = 0.5$ such that it is in a similar scale to $\mathcal{L}_{task}$. 
For the outer optimization to obtain $(\phi'_i,\theta'_i)$, we use Adam optimizer with learning rate $\beta = 5\times10^{-5}$. 
For the inner optimization to update $(\phi,\theta)$, we use plain, non-adaptive gradient descent with learning rate $\alpha = 5\times10^{-5}$, and clip the gradients by norm (threshold by 2.0) to prevent them from exploding.
To construct the meta-task during each iteration, the samples with mini-batch size = 120 are selected from each meta-train and meta-test domains.
The total number of training iterations is 10000.
Our experiments are implemented in Tensorflow with an Nvidia TITAN Xp 12 GB GPU.
On all the experiments, we report the average and standard deviation over three independent runs. 
For liver segmentation, we modify the network backbone as U-Net and reduce the mini-batch size to 60 due to computational overhead. 
The rest settings remain the same. 

\begin{table*}[!t]
	\caption{Quantitative results of DG on epithelium-stroma classification (Acc[\%]).}
	\label{tab:results_hist}
  \resizebox{\linewidth}{!}{
	\begin{tabular}{cc|ccccc|cc}
	  \toprule
	  \multicolumn{1}{c}{\multirow{2}{0.8cm}{Source}}
   &\multicolumn{1}{c|}{\multirow{2}{0.8cm}{Target}}
   &\multicolumn{1}{c}{\multirow{1}{*}{MLDG}}
   &\multicolumn{1}{c}{\multirow{1}{*}{Epi-FCR}}
   &\multicolumn{1}{c}{\multirow{1}{*}{MetaReg}}
   &\multicolumn{1}{c}{\multirow{1}{*}{JiGen}}
   &\multicolumn{1}{c|}{\multirow{1}{*}{MASF}}
   &\multicolumn{1}{c}{\multirow{1}{*}{DeepAll}}
   &\multicolumn{1}{c}{\multirow{1}{*}{ \textbf{ETTA-SE}}}\\
   \multicolumn{1}{c}{} & \multicolumn{1}{c|}{} & \multicolumn{1}{c}{\cite{li2018domain}}& \multicolumn{1}{c}{\cite{li2018learning}}& \multicolumn{1}{c}{\cite{li2019episodic}}& \multicolumn{1}{c}{\cite{jigsaw}}& \multicolumn{1}{c|}{\cite{masf}}& \multicolumn{1}{c}{(Baseline)}& \multicolumn{1}{c}{ \textbf{(Ours})} \\
	  \hline
	  NKI, IHC, NCH & VGH & 91.13$\pm$0.08 & 91.49$\pm$0.13 & 91.74$\pm$0.09 & 92.05$\pm$0.11  & 92.43$\pm$0.13 & 90.43$\pm$0.21 & \textbf{93.51$\pm$0.19}\\
	  VGH, IHC, NCH & NKI & 89.98$\pm$0.05 & 90.15$\pm$0.12 & 90.33$\pm$0.17 & 90.46$\pm$0.09  & 90.32$\pm$0.06 & 89.62$\pm$0.19 & \textbf{91.95$\pm$0.17}\\
	  VGH, NKI, NCH & IHC & 86.68$\pm$0.07 & 87.56$\pm$0.08 & 87.77$\pm$0.19 & 88.29$\pm$0.13  & 88.37$\pm$0.16 & 86.48$\pm$0.08 & \textbf{90.32$\pm$0.12}\\
	  VGH, NKI, IHC & NCH & 87.69$\pm$0.16 & 88.02$\pm$0.31 & 88.57$\pm$0.21 & 89.05$\pm$0.10   & 89.52$\pm$0.06 & 87.17$\pm$0.13 & \textbf{91.28$\pm$0.18}\\
	  \hline
	 \multicolumn{2}{c|}{Average} & 88.85 & 89.31 & 89.60 & 89.96 & 90.16 & 88.43 &  \textbf{91.77}\\
	\bottomrule
  \end{tabular}
  }
\end{table*}

\subsection{Experiments on the epithelium-stroma Classification in Histopathological Images}

\subsubsection{Comparison with Baseline Model}
\begin{figure}[!h]
	\centering
	\subfigure[DeepAll baseline.]
	{
		\begin{minipage}[b]{0.5\linewidth}
			\centering
			\includegraphics[width=\linewidth,height=3.1cm]{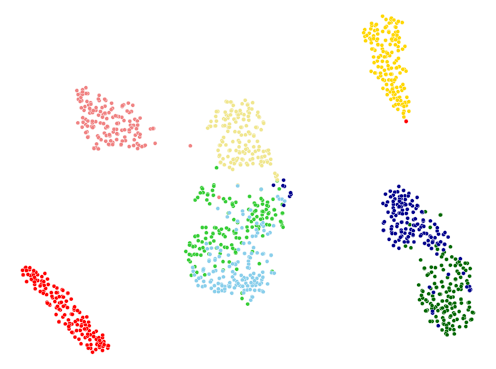}
		\end{minipage}
	}
	\subfigure[Our proposed ETTA-SE.]  
	{
		 \begin{minipage}[b]{0.43\linewidth} 
			\centering
			\includegraphics[width=\linewidth,height=3.1cm]{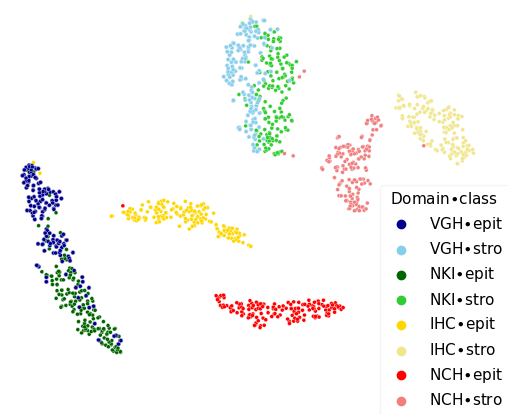}
		\end{minipage}
	}
	\caption{The t-SNE visualization of feature embedding learned by (a) DeepAll baseline and (b) our proposed ETTA-SE. The color type indicates domain ('blue'-unseen test domain; 'green','golden','red'-source domains) and color shade indicates class ('dark'-epithelium class; 'light'-stroma class).}
	\label{fig:tsne} 
\end{figure}

We merge the images from all the source domains and use them to train the DG baseline in a supervised way, as the DeepAll model.
For clarity, the overall proposed method is called as {\itshape episodic training with task augmentation} to regularize the deep {\itshape semantic embedding} (ETTA-SE). 
Table \ref{tab:results_hist} presents the quantitative results on epithelium-stroma classification by them.
It can be observed that the performance of our method is better than the baseline in all the settings, raising the accuracy from 88.43\% to 91.77\% on average.
Moreover, t-SNE \cite{maaten2008visualizing} is used to analyze the extracted features from $\mathcal{F}_\phi$ by DeepAll and our ETTA-SE, as shown in Figure \ref{fig:tsne}. 
The color type (i.e., blue, green, gold, red) represents the domain, and the color shade (i.e., dark, light) denotes the class.
It appears that our method produces a better separation of classes and learns the more domain-invariant embedding, compared to the DeepAll baseline. In particular, for the unseen test domain (blue color),
some samples of epithelium class (dark blue) are pushed apart from the cluster of stroma class (light blue) by our method.

\subsubsection{Comparison with Other DG methods}
We compare with recent state-of-the-art DG methods, which are described as follows.
\textbf{MLDG} \cite{li2018learning}: A meta-learning method for domain generalization, which contains a model-agnostic training procedure to simulates train/test domain shift during training.
\textbf{Epi-FCR} \cite{li2019episodic}: A single deep network with an episodic training procedure, which exposes the model to domain shift.
\textbf{MetaReg} \cite{balaji2018metareg}: A method that uses episodic training with regularization functions to encode the notion of domain generalization.
\textbf{JiGen} \cite{jigsaw}: A model trained by the gradient signals from the supervised learning for semantic labels and the self-supervised learning for a puzzle problem.
\textbf{MASF} \cite{masf}: A recent state-of-the-art method that utilizes the paradigm of model-agnostic episodic learning for semantic features.
These methods are implemented on the task of epithelium-stroma classification using the public codes provided by their authors. 
From the quantitative results list in Table \ref{tab:results_hist}, it is observed that MLDG improves slightly compared to DeepAll baseline.
Epi-FCR and MetaReg further enhance the performance. The not bad results of MLDG, Epi-FCR, MetaReg indicate that exposing the model to domain shift simulation episodically benefits the generalization performance of deep models to unseen domains.
JiGen and MASF improve 1.53\% and 1.73\% at average accuracy compared to the DeepAll baseline. The performance achieved by MASF shows the effectiveness of introducing the episodic training scheme in the learning of semantic features.
Finally, the proposed ETTA-SE, which incorporates the task augmentation as well as deep embedded regularization, outperforms the state-of-the-art DG methods on 
 Across all the settings on different target domains, it appears that our method consistently outperforms the state-of-the-art DG methods.

\begin{table*}
	\caption{Ablation study on key components of our method on epithelium-stroma classification (Acc[\%]).}
	\label{tab:abla_hist}
	\centering	
	\resizebox{\linewidth}{!}{
	\setlength{\tabcolsep}{2mm}{
	\begin{tabular}{cccc|cccc|c}
	  \toprule
	ET& TA &$\mathcal{L}_{sa}$ & $\mathcal{L}_{pa}$ &NKI,IHC,NCH-VGH   &VGH,IHC,NCH-NKI    &VGH,NKI,NCH-IHC    &VGH,NKI,IHC-IHC      &Average\\

	\hline
	  -  &-                           &-         &-         & 90.43$\pm$0.21 & 89.62$\pm$0.19 & 86.48$\pm$0.08 & 87.17$\pm$0.13 & 88.43\\
	  \hline
  
	  \checkmark &-                            &-         &-         & 91.27$\pm$0.23 & 90.67$\pm$0.15 & 86.17$\pm$0.25 & 89.81$\pm$0.09 & 89.48\\
	  \checkmark& \checkmark                          &-         &-         & 91.33$\pm$0.27 & 90.97$\pm$0.18 & 88.76$\pm$0.14 & 90.90$\pm$0.10 & 90.49\\
	  \checkmark &\checkmark   &\checkmark   &-         & 91.96$\pm$0.24 & 90.93$\pm$0.28 & 89.84$\pm$0.10 & 91.18$\pm$0.19 & 90.98\\
	  \checkmark &\checkmark&   -&\checkmark   & 92.78$\pm$0.19 & 91.15$\pm$0.24 & 90.12$\pm$0.13 & 91.16$\pm$0.24 & 91.27\\
	  \hline
	  \checkmark &\checkmark  &\checkmark & \checkmark    & 93.51$\pm$0.19 & 91.95$\pm$0.17 & 90.32$\pm$0.12 & 91.28$\pm$0.18 & 91.77\\
	\bottomrule
  \end{tabular}
  }
	}
\end{table*}


\subsubsection{Ablation Study}
We conduct an ablation study to investigate five key issues in our ETTA-SE framework on the task of epithelium-stroma classification : 1) contribution of each component, 2) effect of task augmentation (TA) against task-level overfitting, 3) ablation on the mixed degree of mixed task sampling (MTS) strategy,
4) impact of the domain alignment of semantic embedding (SE) between source and unseen domains,
5) stability analysis of our overall model.

\textbf{Contribution of each component.} We validate the effect of the four key components in our method, i.e., episodic training (ET), task augmentation (TA), the loss of sample-wise alignment $\mathcal{L}_{sa}$ and prototype-wise alignment $\mathcal{L}_{pa}$, respectively, by an ablation study shown in Table \ref{tab:abla_hist}. 
The 1st row corresponds to the DeepAll baseline, where the standard training is performed by merging all the data from source domains. 
The last row corresponds to the proposed ETTA-SE.
Note that when neither $\mathcal{L}_{fa}$ nor $\mathcal{L}_{ca}$ is added (i.e., ETTA, without SE), a task loss calculated on meta-test is used as the basic $\mathcal{L}_{meta}$.
It can be observed that each component plays its own role. 
Specifically, compared to DeepAll,
ET improves 1.05\% at average Acc, showing the effect of simulating domain shift to solve DG (2nd row). TA further improves 1.01\%, which demonstrates it is necessary to introduce task augmentation (3rd row). 
Meanwhile, the two loss terms for regularizing deep embedded alignment are both beneficial to the model performance. 
More importantly, we observe that the performance gain achieved by using them together is greater than the sum of the gain by using them respectively. 
Thus, these two perspectives for deep embedded alignment appear to be complementary to each other.

\begin{figure}[htbp]
	\centering
	\subfigure[ET w/o TA]    
	{
		 \begin{minipage}[b]{0.44\linewidth}   
			\centering
			\includegraphics[width=\linewidth,height=3.2cm]{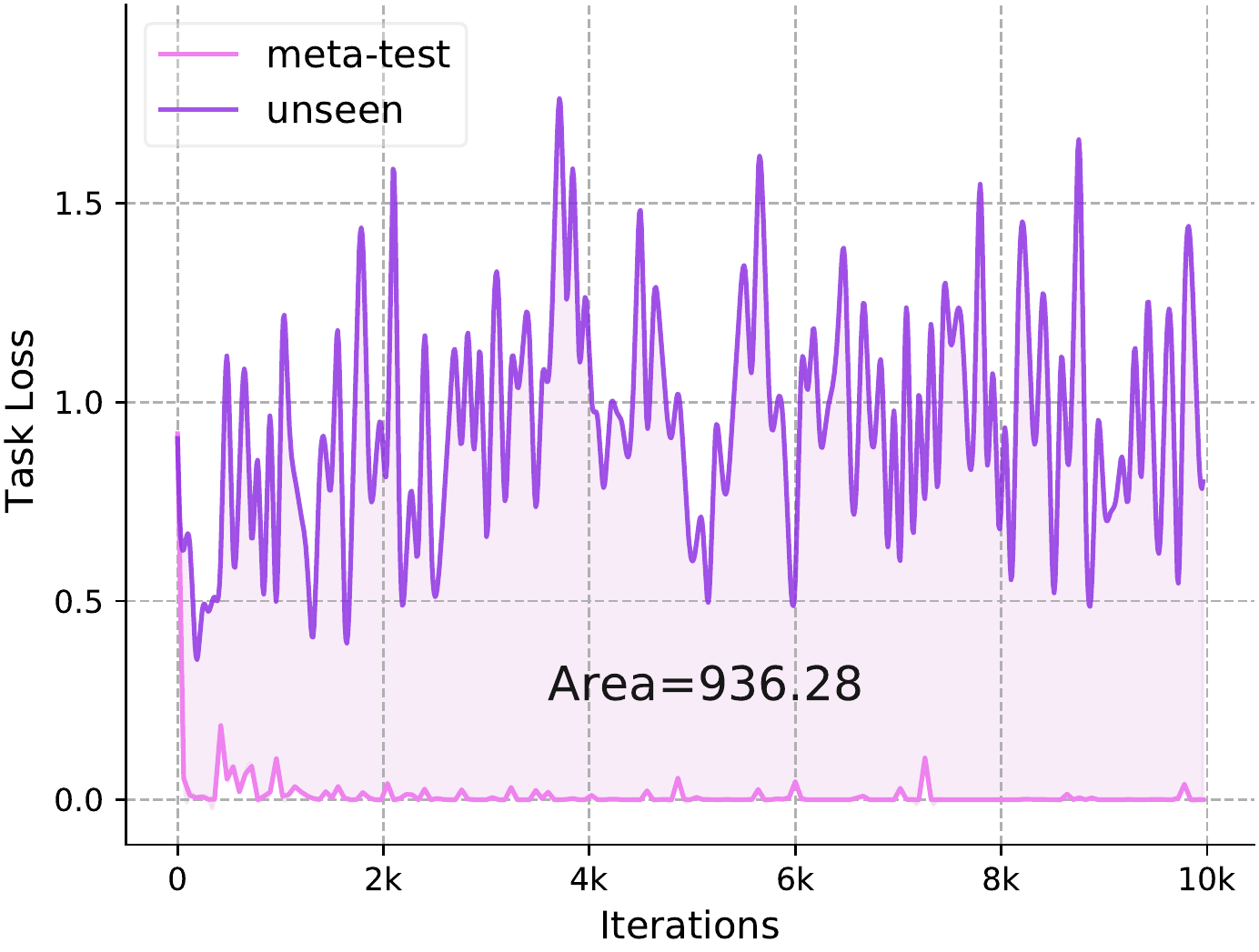} 
		\end{minipage} 
	}
	\subfigure[ET w/ TA]
	{
		\begin{minipage}[b]{0.44\linewidth}   
			\centering
			\includegraphics[width=\linewidth,height=3.2cm]{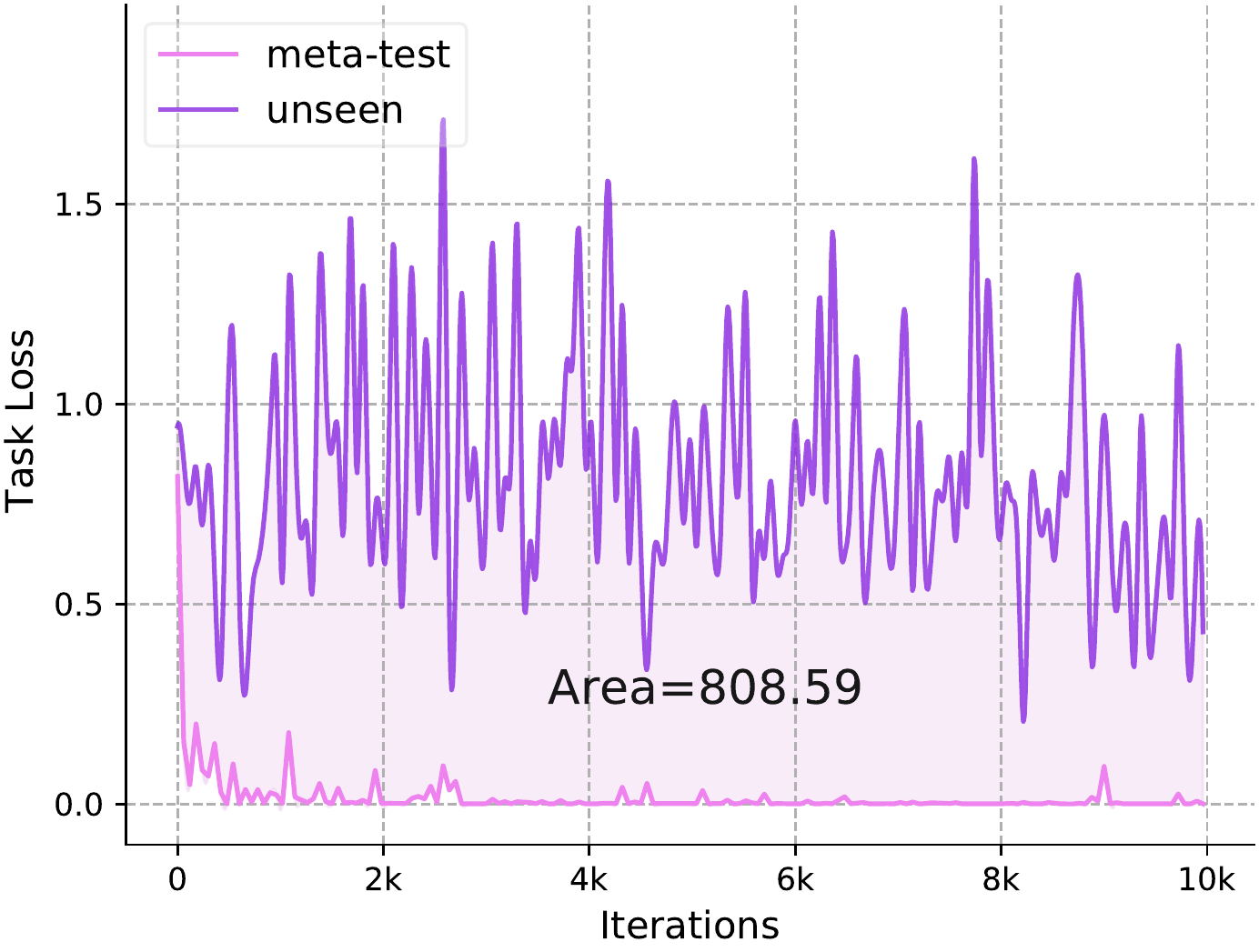} 
		\end{minipage}       
	}
	\caption{The exhibition of the task-level overfitting.
	The area of shaded region is calculated to measure the performance gap between training and testing tasks (i.e., overfitting degree).
	'ET'-episodic training, 'TA'-task augmentation.
	}
	\label{fig:loss_hist}
\end{figure}

\textbf{Analysis of task augmentation against overfitting.} 
We conduct the ablation analysis on the paradigm of episodic training (ET) w/ and w/o our task augmentation (TA).
In standard supervised learning, a common approach to observe overfitting is by comparing the task loss on the training and testing sets. 
Accordingly, in episodic training, we observe the loss of DG performance on the training and testing tasks, respectively.
Concretely, we use the task loss on the meta-test domain as the former and that on the unseen test domain as the latter.
As shown in Figure \ref{fig:loss_hist}, the model can fit to the training tasks well during ET whether with TA or not. 
However, the obvious gap (shaded region) between the loss curves exhibits the overfitting phenomenon to training tasks.
We further calculate the area of shaded region as an approximate observation for overfitting degree.
When equipping the ET paradigm with TA,
the area decreases from 936.28 to 808.59, which shows that the mitigation of the task-level overfitting is achieved by the proposed TA.



\begin{figure}[htbp]
	\centering
	\begin{minipage}[t]{0.47\columnwidth}
	\centering
	\includegraphics[width=\linewidth,height=3cm]{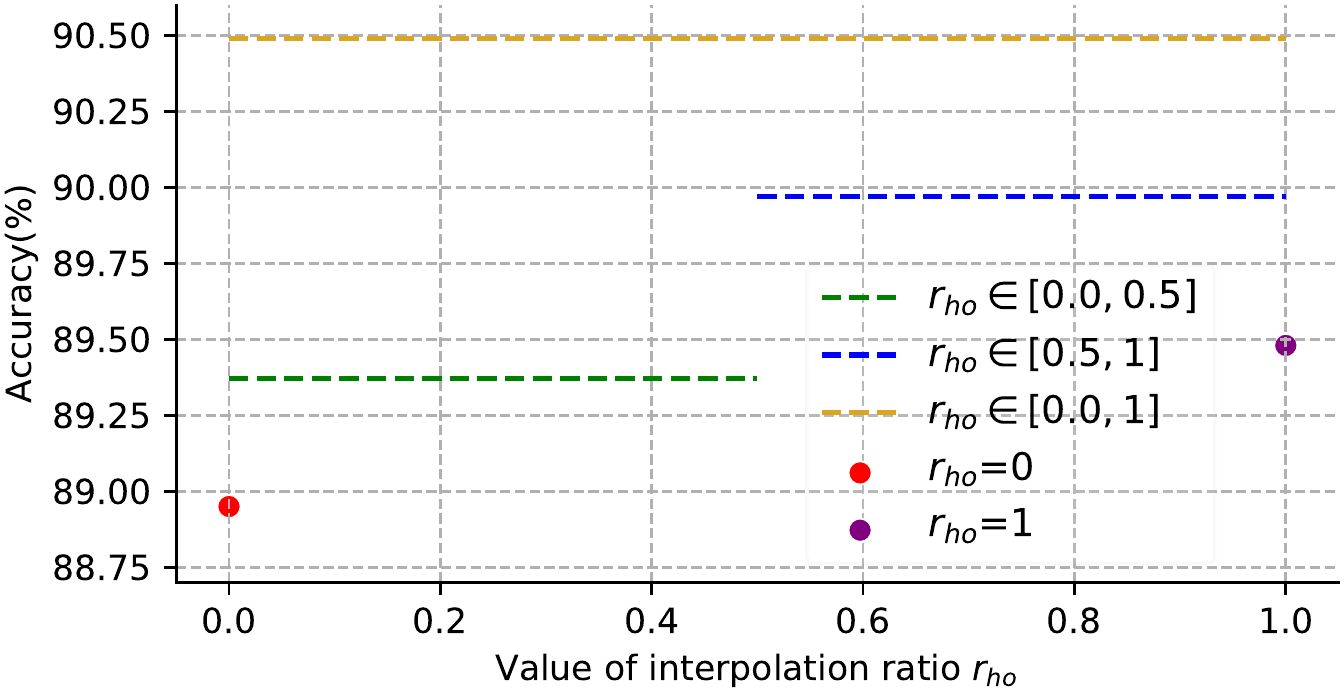}
	\caption{Ablation on the mixed degree of MTS.}
	\label{fig:conti}
	\end{minipage}
	\begin{minipage}[t]{0.43\columnwidth}
	\centering
	\includegraphics[width=\linewidth,height=2.88cm]{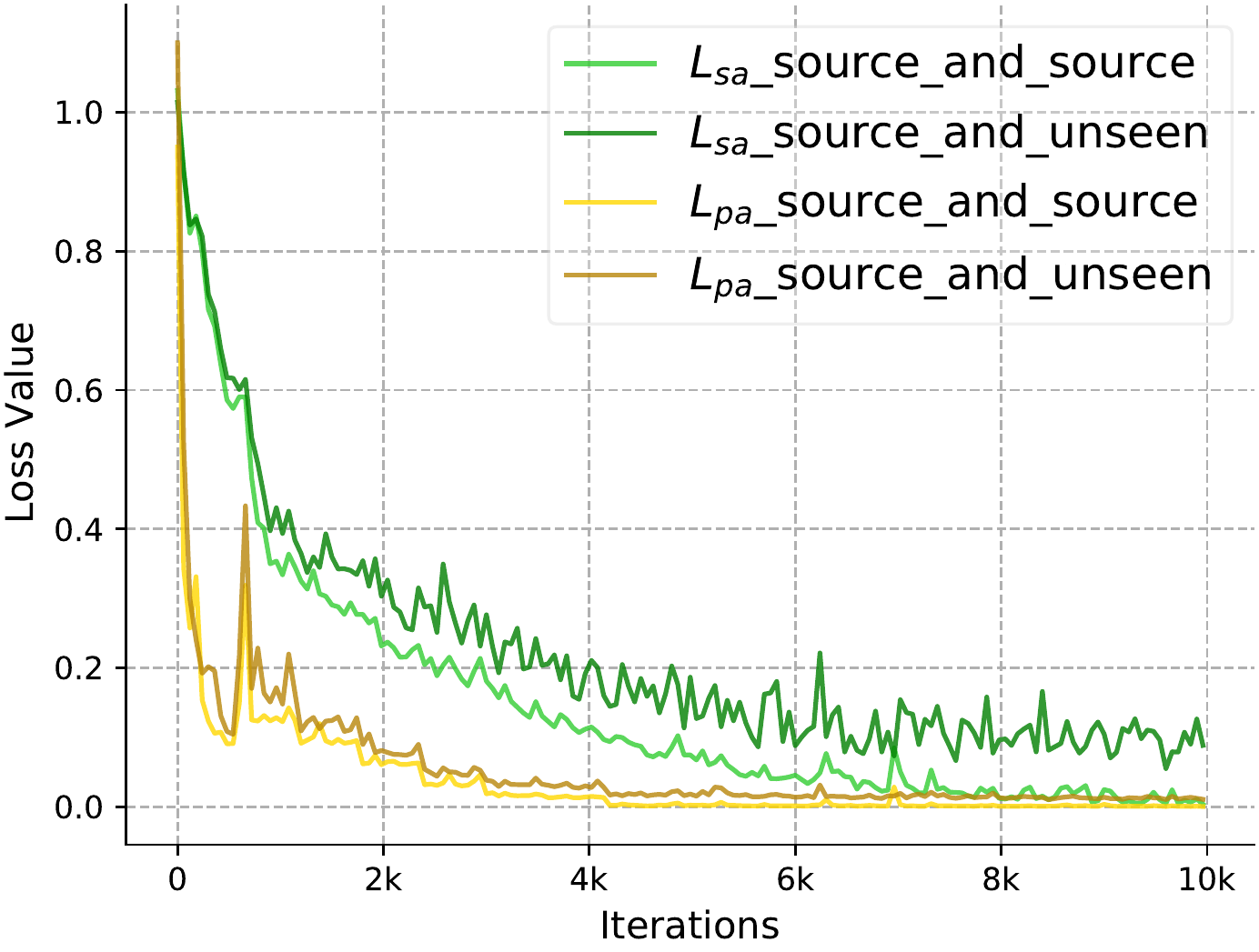}
	\caption{Effect of the domain alignment on SE.}
	\label{fig:semantic_loss}
	\end{minipage}
	\end{figure} 
	
\textbf{Ablation on the mixed degree of mixed task sampling.} 
We investigate how the enhanced variety by task augmentation affects the performance of DG, by an ablation study on the mixed degree of mixed task.
Specifically,
given $K$ source domains, $K-1$ domains are randomly selected as meta-train, and the held-out domain denote as (denoted as $\mathcal{D}_{ho}$) as meta-test, using the task sampling (TS) strategy in existing episodic training paradigm.
In comparison, in the proposed mixed task sampling (MTS), the meta-test domain is acquired by the interpolation from all K source domain (cf. Eq. \ref{eq_Dte}). 
To facilitate illustration, we denote the interpolation ratio of held-out domain $\mathcal{D}_{ho}$ as $r_{ho}$.
As shown in Figure \ref{fig:conti},
we conduct experiments of MTS with $r_{ho}$ in different situations, including dynamic values with the range of $[0.0, 0.5]$, $[0.5, 0.1]$, $[0.0, 1.0]$, and fixed values at $0.0$, $1.0$.   
$r_{ho}=0$ corresponds to $\mathcal{D}_{ho}$ excluded in the interpolation process for meta-test generation, which means the meta-test domain is just the combination of meta-train domains.
It appears to behave not well due to the signal of domain shift simulation is too weak in such a  situation.
$r_{ho}=1$ corresponds to TS and seems to outperforms $r_{ho}=0$, which shows the effectivenss of episodic simulation for virtual training tasks of DG. 
However, in TS, the risk of overfitting to virtual training tasks prevents the model from achieving the good performance on the real testing DG task.
As a comparison, we observe the proposed MTS outperforms TS under most variations of $r_{ho}$, which demonstrates the validity of enhancing variety during the meta-task sampling process.
An exception is $r_{ho}\in[0,0.5]$ that achieves the lower but very color results with TS, which is probably due to the still weak domain shift simulation. 
Finally, we see the proposed MTS yields the best results under $r_{ho}\in[0,1]$.

\textbf{Semantic embedded alignment between source and unseen domains.}
The proposed meta-objective performs the alignment regularization on the semantic embedding across source domains episodically.
Here we analyze the domain alignment of the deep embedding between source and unseen test domains, which reflects the effect of meta-objective on the real DG task. 
As shown in Figure \ref{fig:semantic_loss}, the loss curves of $L_{sa}$ and $L_{pa}$ are drawn to exhibit the effect of domain alignment.
We see the alignment losses between source domains (light-colored curves) falls normally during the training phase.
At the same time, 
the losses between source and unseen domains (dark-colored curves) drops following the trends of those between source domains.
This demonstrates although the proposed meta-objective for domain alignment is applied on the semantic embedding between source domains,
the regularization effect can be generalized to the real DG task, which promotes the deep embedded alignment between source and unseen test domains.


\begin{table*}[!t]
	\caption{Quantitative results of DG on liver segmentation (Dice[\%]).}
	\centering
	\label{tab:results_ct}
	\resizebox{\linewidth}{!}{ 
	\begin{tabular}{cc|ccccc|cc}
		\toprule
		\multicolumn{1}{c}{\multirow{2}{0.8cm}{Source}}
	&\multicolumn{1}{c|}{\multirow{2}{0.8cm}{Target}}
	&\multicolumn{1}{c}{\multirow{1}{*}{MLDG}}
	&\multicolumn{1}{c}{\multirow{1}{*}{Epi-FCR}}
	&\multicolumn{1}{c}{\multirow{1}{*}{MetaReg}}
	&\multicolumn{1}{c}{\multirow{1}{*}{JiGen}}
	&\multicolumn{1}{c|}{\multirow{1}{*}{MASF}}
	&\multicolumn{1}{c}{\multirow{1}{*}{DeepAll}}
	&\multicolumn{1}{c}{\multirow{1}{*}{ \textbf{ETTA-SE}}}\\
	\multicolumn{1}{c}{} & \multicolumn{1}{c|}{} & \multicolumn{1}{c}{\cite{li2018domain}}& \multicolumn{1}{c}{\cite{li2018learning}}& \multicolumn{1}{c}{\cite{li2019episodic}}& \multicolumn{1}{c}{\cite{jigsaw}}& \multicolumn{1}{c|}{\cite{masf}}& \multicolumn{1}{c}{(Baseline)}& \multicolumn{1}{c}{ \textbf{(Ours)}} \\
		\hline
		CHAOS, IRCAD, LITS & BTCV  & 80.40$\pm$0.24 & 82.50$\pm$0.11 & 84.83$\pm$0.23 & 85.90$\pm$0.20  & 86.29$\pm$0.23 & 81.26$\pm$0.18 &  \textbf{86.70$\pm$0.24}\\
		BTCV, IRCAD, LITS  & CHAOS & 87.22$\pm$0.14 & 90.42$\pm$0.08 & 91.08$\pm$0.19 & 90.93$\pm$0.16  & 91.18$\pm$0.16 & 84.38$\pm$0.15 &  \textbf{91.93$\pm$0.19}\\
		BTCV, CHAOS, LITS  & IRCAD & 89.17$\pm$0.12 & 89.26$\pm$0.13 & 89.17$\pm$0.09 & 91.44$\pm$0.24  & 90.89$\pm$0.10 & 90.10$\pm$0.13 &  \textbf{92.14$\pm$0.09}\\
		BTCV, CHAOS, IRCAD & LITS  & 88.67$\pm$0.08 & 88.48$\pm$0.21 & 89.41$\pm$0.15 & 89.13$\pm$0.21  &  \textbf{90.11$\pm$0.18} & 88.19$\pm$0.24 & 89.73$\pm$0.26\\
		\hline
		\multicolumn{2}{c|}{Average}& 86.37          & 87.67           & 88.63         & 89.16           & 89.54          & 85.98          &  \textbf{90.20}\\
	\bottomrule
	\end{tabular}
	}
\end{table*}  

\begin{figure}[htbp]
	\centering
	\begin{minipage}{0.4\columnwidth}
	\centering
	\includegraphics[width=0.95\linewidth,height=2.75cm]{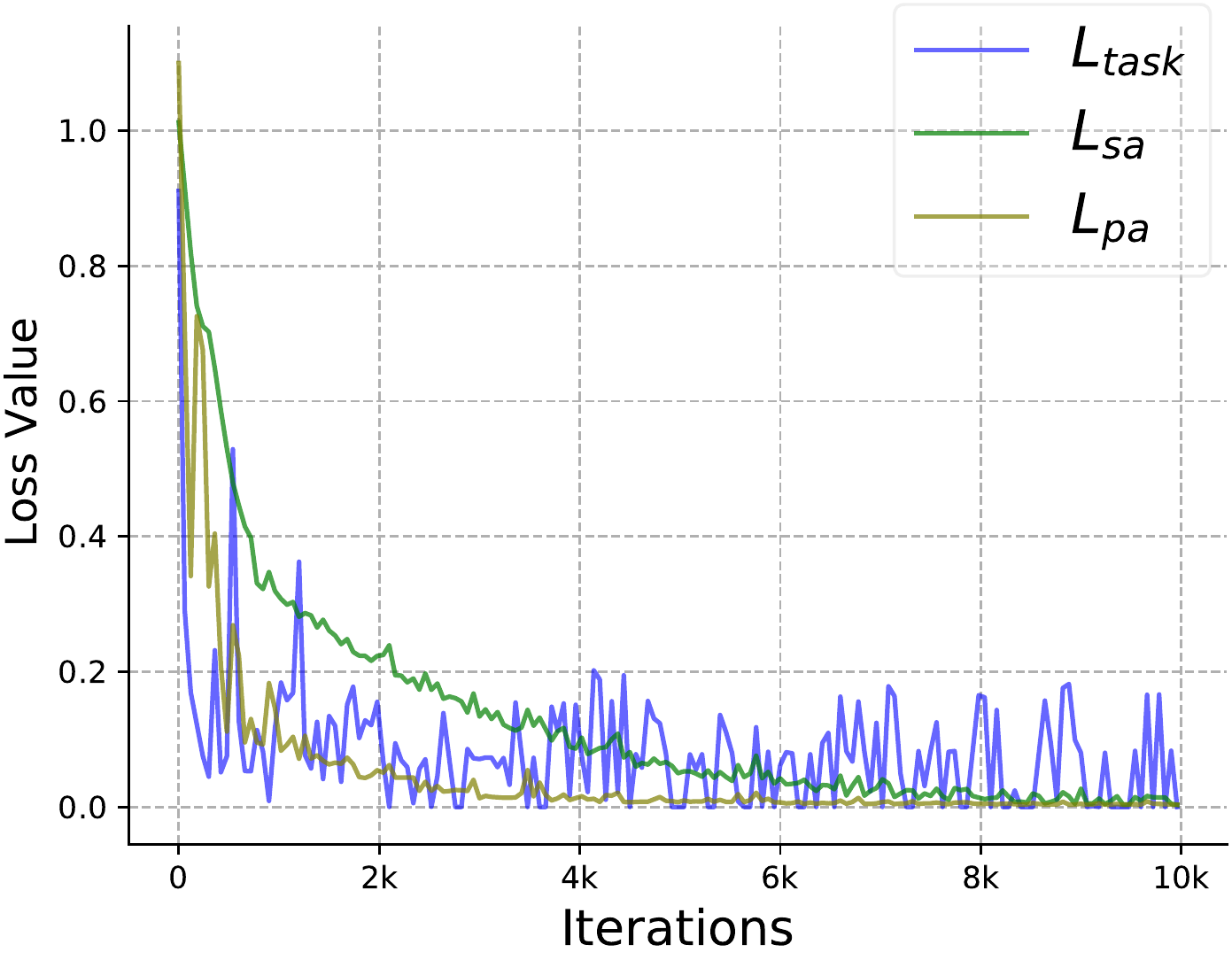}
	\caption{Training Loss.}
	\label{fig:stable_loss} 
	\end{minipage}
	\begin{minipage}{0.58\columnwidth}
	\centering
	\subfigure
	{
		\begin{minipage}{\linewidth}  
			\centering
			\includegraphics[width=0.9\linewidth,height=1.1cm]{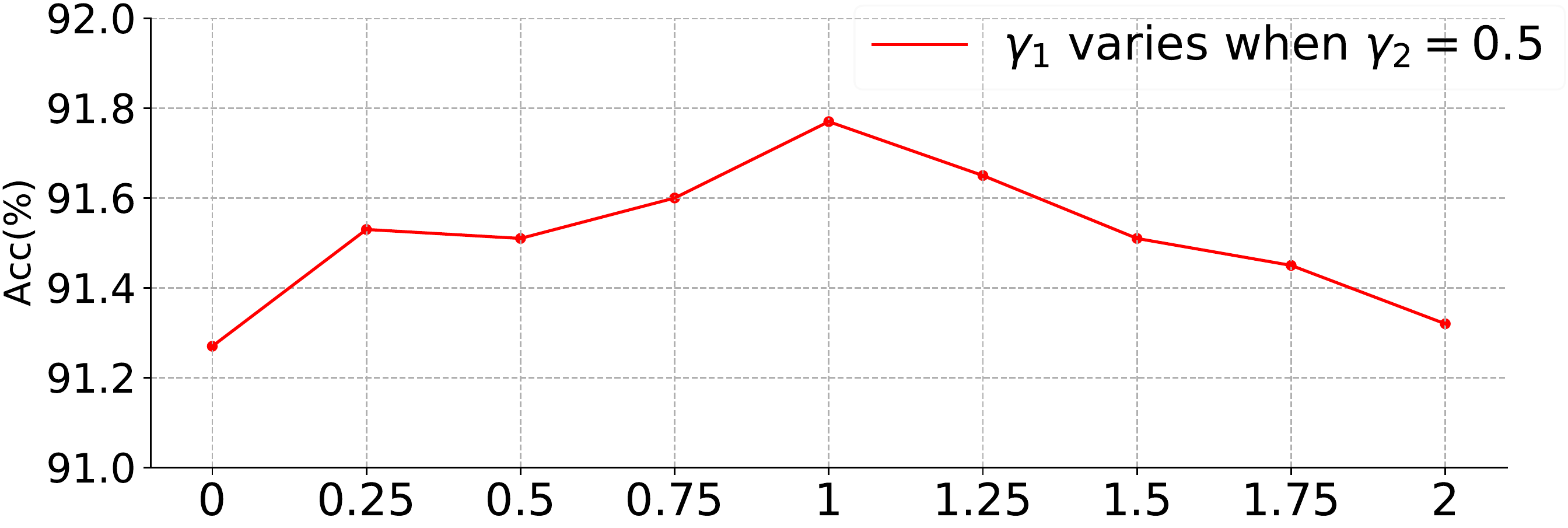}
		\end{minipage}
	}\\ 
	\subfigure
	{
		 \begin{minipage}{\linewidth} 
			\centering
			\includegraphics[width=0.9\linewidth,height=1.1cm]{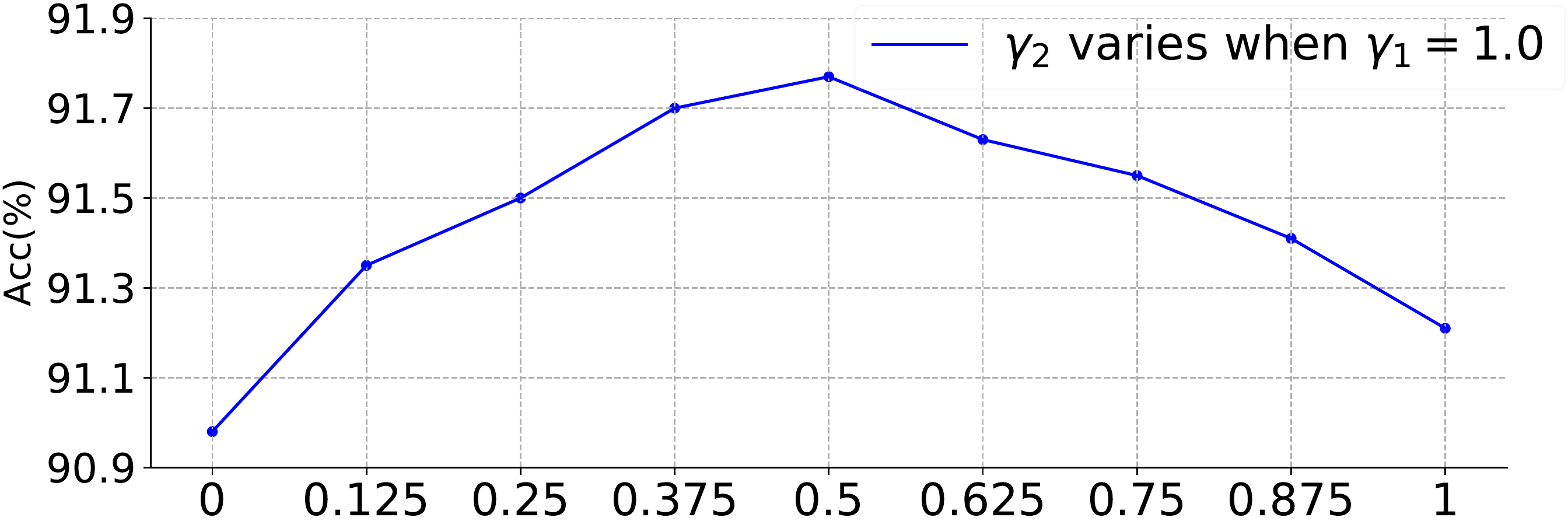}
		\end{minipage}
	}
	\caption{Sensitivity study on$\gamma_1$, $\gamma_2$.}
	\label{fig:gamma}
	\end{minipage}
	\end{figure}

\textbf{Training stability and parameter sensitivity.} 
To evaluate the training stability of the proposed ETTA-SE framework, the curves of the loss terms involved in the training procedure are all drawn in Figure \ref{fig:stable_loss}.
We observe that they drop in a systematic way as the training goes, which shows the stability of our system.
We also see that $\mathcal{L}_{task}$ decreases in the early training stage and then oscillates around low values. 
It demonstrates that the meta-objective for deep embedded alignment, i.e., $\mathcal{L}_{fa}$ and $\mathcal{L}_{ca}$, does perform the regularization effect on the model.
That is, the model is promoted to perform the appropriately well rather than extremely well on the training set, which is more in line with the goal of generalizing to unseen domains.  

In addition, we evaluate the sensitivity of our ETTA-SE with respect to the trade-off parameters of meta-objective, i.e., $\gamma_1$ and $\gamma_2$. Figure \ref{fig:gamma} shows the performance of our proposed model when $\gamma_1$ ($\gamma_2$) varies and the other parameter $\gamma_2$ ($\gamma_1$) is fixed. It appears that the performance of our model is not sensitive to the variations of $\gamma_1$ ($\gamma_2$) especially when $\gamma_1$ ($\gamma_2$) varies from 0.25 (0.25) to 1.5 (0.75).

\begin{figure*}[!t]
	\centering
	\includegraphics[width=\textwidth,keepaspectratio]{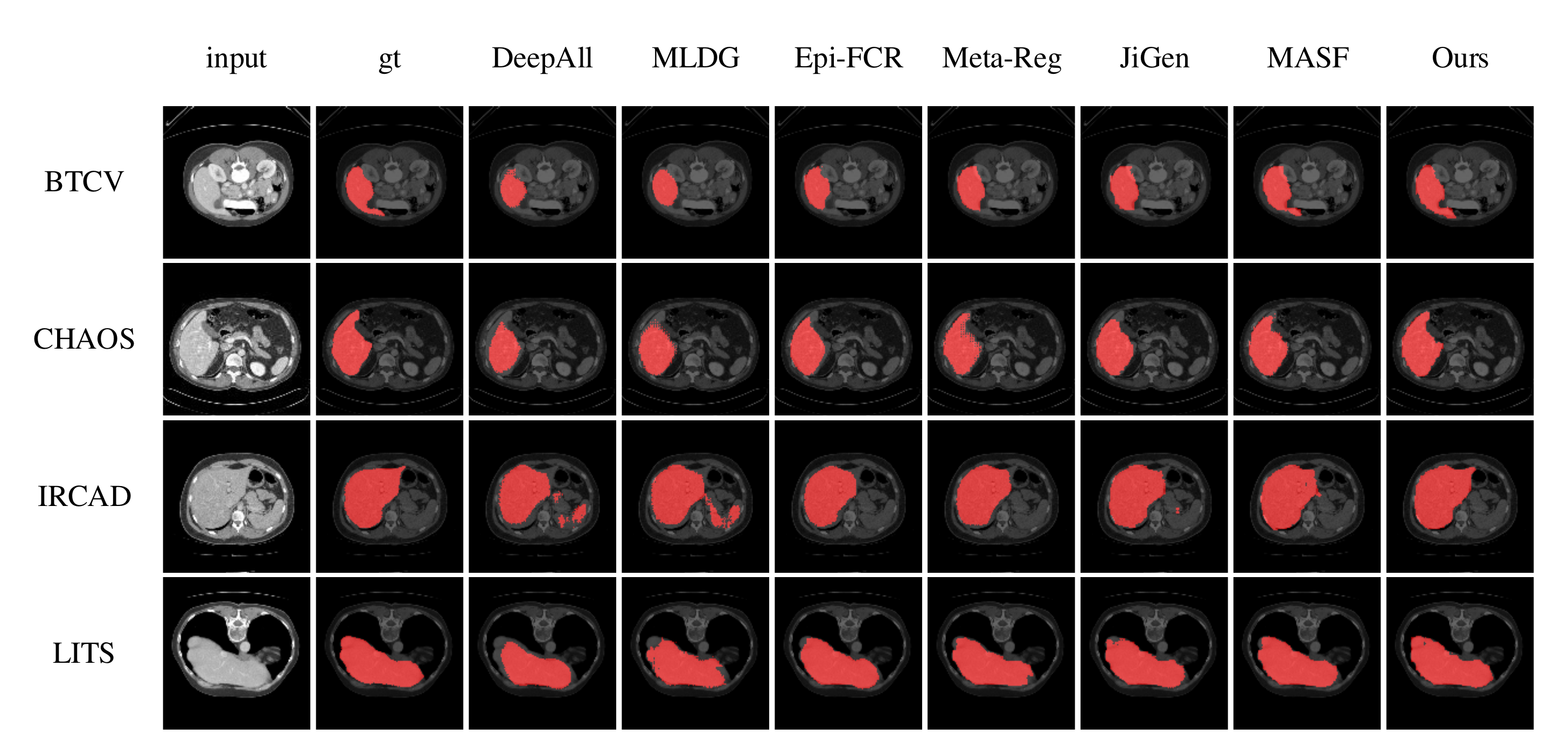} 
	\caption{Visual results of DG on liver segmentation.}
	\label{fig:liver_seg}
\end{figure*}

\subsection{Experiments on the liver segmentation in CT images}

\subsubsection{Comparison with Baseline Model}
We also evaluate our DG method on the liver segmentation in CT images. 
Table \ref{tab:results_ct} presents the results of the DeepAll baseline and our proposed ETTA-SE method. 
Although the domain shift is not easy to observe directly from the grayscale CT scans, it makes the cross-domain performance of liver segmentation drops a lot. 
In addition, the degree of performance degradation varies under different unseen test domains, which shows the diverse difficulties of these DG missions.
Compared to the DeepAll baseline, our proposed ETTA-SE achieves the improvement as 4.22\% Dice score on average. 
As shown in Figure \ref{fig:liver_seg}, the DeepAll baseline fails to accurately segment the liver regions in some cases.
In comparison, the results yielded by our method are more refined, and exhibit the greater visual similarity to the ground-truth.  

\subsubsection{Comparison with Other DG methods}
We further report the quantitative results compared with the prior DG methods, as shown in Table \ref{tab:results_hist}.
We implement these approaches on the task of liver segmentation based on the official public codes with some necessary modifications.
When the source domains are BTCV, CHAOS, IRCAD and the unseen test domain is LITS, the proposed ETTA-SE method does not outperforms MASF \cite{masf}.
However, our method can still achieve the state-of-the-art performance on all the remaining settings.
The proposed method also surpasses other advanced methods on the average Dice score. 
As shown in Figure \ref{fig:liver_seg}, some over-segmented and under-segmented results produced by prior methods can be refined by our method.   

\section{Conclusion}
In this paper, we propose the novel episodic training with task augmentation method for domain generalization (DG) on medical imaging classification.
The episodic training paradigm transfers the knowledge of DG from meta-task simulation to the real task of DG.
Inspired by the limited number of available source domains in clinical practice, we consider the special risk of overfitting to training meta-tasks.
To alleviate it, task augmentation is developed to enhance the variety during training task generation, using mixed task sampling (MTS).
The meta-objective is integrated into the established learning framework to regularize the deep semantic embedding to align across training domains.
The results of extensive experiments demonstrate our state-of-the-art performance for DG. 

\footnotesize
\bibliographystyle{ieeetr}
\bibliography{ref}

\end{document}